\documentclass[12pt]{article}
\usepackage{times}  % DO NOT CHANGE THIS
\usepackage{helvet}  % DO NOT CHANGE THIS
\usepackage{courier}  % DO NOT CHANGE THIS
\usepackage[hyphens]{url}  % DO NOT CHANGE THIS
\usepackage{graphicx} % DO NOT CHANGE THIS
\usepackage{caption} % DO NOT CHANGE THIS AND DO NOT ADD ANY OPTIONS TO IT
\usepackage{algorithm}
\usepackage{algorithmic}
\usepackage{subcaption} % needed for subfigures
\usepackage{bm}
\usepackage{amsmath}
\usepackage{amsfonts}
\usepackage{amssymb}
\usepackage{xcolor}
\usepackage{amsthm}

\usepackage{stmaryrd}
\usepackage{enumitem}
\usepackage{array}

\usepackage{newfloat}
\usepackage{listings}
\DeclareCaptionStyle{ruled}{labelfont=normalfont,labelsep=colon,strut=off} % DO NOT CHANGE THIS
\floatstyle{ruled}
\newfloat{listing}{tb}{lst}{}
\floatname{listing}{Listing}
\usepackage[linktoc=all]{hyperref}
\hypersetup{colorlinks=true,
	linkcolor=blue,
	citecolor=blue
}

\usepackage[margin=1in]{geometry}
\usepackage{titlesec}

\titleformat{\paragraph}
{\normalfont\normalsize\bfseries}{\theparagraph}{1em}{}
\titlespacing*{\paragraph}
{0pt}{3.25ex plus 1ex minus .2ex}{1.5ex plus .2ex}

\title{A Deep Learning Framework for Solving Hyperbolic Partial Differential Equations: Part I}
\author{
    Rajat Arora\thanks{Senior member of technical staff at Advanced Micro Devices, Inc. (AMD).}
}
\newsavebox{\largestimage}

\begin{document}

\newcommand{\beginmyitemize} {\begin{itemize} [leftmargin=1.5em, itemsep=1pt,topsep=1pt,parsep=1pt,partopsep=0pt]} %leftmargin=2em,
	
\pagenumbering{arabic}
\date{}% not using this gives today's date

\maketitle

\begin{abstract}

% \textcolor{red}{ADD WEAK FORM}
% ,  related to the collocation-based PDE discretization.

Physics informed neural networks (PINNs) have emerged as a powerful tool to provide robust and accurate approximations of solutions to partial differential equations (PDEs). However, PINNs face serious difficulties and challenges when trying to approximate PDEs with dominant hyperbolic character. This research focuses on the development of a physics informed deep learning framework to approximate solutions to nonlinear PDEs that can develop shocks or discontinuities without any \textit{a-priori} knowledge of the solution or the location of the discontinuities. The work takes motivation from finite element method that solves for solution values at nodes in the discretized domain and use these nodal values to obtain a globally defined solution field. Built on the rigorous mathematical foundations of the discontinuous Galerkin method, the framework naturally handles imposition of boundary conditions (Neumann/Dirichlet), entropy conditions, and regularity requirements. Several numerical experiments and validation with analytical solutions demonstrate the accuracy, robustness, and effectiveness of the proposed framework.

 % Prior works on using machine learning to approximate solutions to hyperbolic PDEs 

% Physics-informed neural network architectures have emerged as a powerful tool for developing flexible PDE solvers that easily assimilate data. When applied to problems in shock physics however, these approaches face challenges related to the collocation-based PDE discretization underpinning them. By instead adopting a least squares space-time control volume scheme, we obtain a scheme which more naturally handles: regularity requirements, imposition of boundary conditions, entropy compatibility, and conservation, substantially reducing requisite hyperparameters in the process.

% Therefore, hybrid strategies integrating physics-informed ML with traditional approaches are emerging as a promising option to tackle this computational challenge of solving complex multi-physics PDEs 

% greatly accelerate the process of scientific investigation and engineering design.

\end{abstract}

% Specifically, we consider a simplified model of elastodynamics on a two-dimensional homogeneous domain.
% Our prior work has demonstrated the capability of PhySRNet to enhance spatial resolution of deformation fields in hyperelastic solids.
% Existing numerical schemes such as finite element methods have been successfully used to predict dynamical behaviour of large scale structural systems.
% However, modelling such large scale systems using numerical methods can be computationally expensive constraining the mesh refinement.
% Prior work on data driven deep learning based image super-resolution models have been proposed to enhance the spatio-temporal resolution of the solution fields of various PDEs obtained on coarse mesh. 
% However, the approach has requires a large amount of high temporal-spatial resolution data to train such models

% is bringing these numerical solvers to their limits.
% the pursuit of high-fidelity multi-physics coupled simulation has 
% led to the ever-increasing sophistication of mathematical models to capture the   
%  number of attempts are being made

%  including, but not limited to, fluid mechanics, solid mechanics, chemical engineering, and biology. 

\section{Introduction}
Partial Differential Equations (PDEs) are at the heart of modeling complex spatio-temporal dynamic systems ubiquitous in many scientific disciplines. Of particular interest in this work are a special class of time-dependent PDEs, called hyperbolic partial differential equations. Hyperbolic PDEs play an important role in various applications in natural sciences and engineering ranging from fluid dynamics \cite{anderson1995computational}, solid mechanics \cite{bonet2015first,achenbach2012wave, arora2019computational} to problems in traffic flow \cite{colombo2003hyperbolic}, acoustics \cite{tam1993dispersion}, and gas dynamics \cite{temple1981solutions}. An important distinguishing feature of  hyperbolic PDEs is that solutions may easily develop shocks or discontinuities even if the initial data is smooth \cite{dafermos2005hyperbolic},  making them mathematically and computationally challenging to solve. Traditional techniques such as finite element/difference/volume \cite{cockburn2012discontinuous, sod1978survey, leveque2002finite} based approaches have been shown to be successful in solving hyperbolic PDEs through the use of suitable upwind discretizations \cite{osher1982upwind}, slope limiters \cite{sweby1984high}, or addition of artificial viscosity. However, huge computational expense still remains a critical issue for solving large system sizes, inverse problems, or assimilating experimental data. 

% to obtain total variation diminishing solutions

% Hyper-bolic PDEs are challenging to solve numerically using classical discretization schemes, because they tend to form self-sharpening, highly-localized, nonlinear shock waves that require specific approximation strategies and fine meshes44 .

% \textcolor{red}{one more line about discontinuity resolving upwinding.}

% Hence, accurate solving of these PDEs is of utmost importance. 

% are ubiquitous in various bracnhes of science 
% Hyperbolic PDEs importance and approaches that are used to solve them.

% However, little attention has been paid to solving complex hyperbolic PDEs with discontinuous solutions or initial conditions. 

% hyperbolic

% Several works have confirmed that the plain-vanilla PINN models face serious challenges while solving PDEs with dominant hyperbolic character.

Alternatively, recent advances in deep learning have led to the development of several data-driven and Physics-Informed Neural Network (PINN) models to solve nonlinear PDEs for both forward and inverse problems \cite{arora2022physics, arora2022spatio, jagtap2022physics}. While PINNs as a PDE solver have shown great strengths in solving multi-dimensional forward and inverse problems, but like any other computational method, it does have some limitations.

Several works have confirmed that PINNs face serious difficulties when trying to approximate solutions with sharp gradients or discontinuities, which is common with hyperbolic PDEs. For example, Fuks et al.  \cite{fuks2020limitations} demonstrated that PINNs fail to provide reasonable approximations to solution to nonlinear hyperbolic PDEs in the absence of any artificial viscosity. Moreover, the quality of the solution, model convergence rate, and the loss landscape highly  depends on the choice of the viscosity parameter.  Mishra et. al \cite{mishra2022estimates} demonstrated that PINNs exhibited poor accuracy while trying to approximate solutions to inviscid scalar conservations laws resulting in large generalization errors. The failure of the governing PDEs to hold in the classical sense in the regions of discontinuities \cite{dafermos2005hyperbolic} further adds to the convergence woes of the current deep-learning based approaches. Moreover, PINNs might converge to unphysical solutions \cite[Sec. 5.1]{patel2022thermodynamically} as the weak solutions to PDEs with dominant hyperbolic character may not be unique. Therefore, the objective function must be augmented with effective entropy conditions and other physical limitations that guarantee uniqueness. Another major limitation of several of the current approaches is that they require \textit{a priori} knowledge of the shock location to predict solutions with sharp gradients \cite{mao2020physics,jagtap2021extended, dwivedi2019distributed, jagtap2020conservative}. When the PDE contains second- or higher-order spatial derivatives, the development of physics informed machine learning model may further suffer from degraded accuracy or convergence issues \cite{arora2022physics}.  A brief review of prior work on using PINNs to solve hyperbolic PDEs and circumventing these issues is presented in Section \ref{sec:prior_works}.

We emphasize that this shortcoming of PINNs to approximate discontinuous functions or solutions to hyperbolic PDEs exists independently of the model architecture or choice of hyper-parameters (number of collocation points, scaling coefficients for different loss components). This is because the approximations of the current neural network architectures reside in the (nonlinear) continuous space which pose challenges to problems with reduced regularity requirements such as those in hyperbolic and $H(div)$/$H(curl)$ problems. Besides, there are no existing theoretical works that can guarantee neural networks to approximate discontinuous functions.   Therefore, we believe that the current deep-learning based frameworks lack the specific ingredients necessary to model the self-sharpening highly-localized, nonlinear shock waves.

To this end, this research presents a Discontinuous Galerkin (DG) based deep-learning framework that is capable of capturing any shocks or discontinuities in the solution field without using any a-priori knowledge of the shock location. The key idea behind using DG is that the solution is discontinuous at the global scale but smooth and continuous at the discrete level (inside elements). This assimilation of DG methodology in the PINNs framework gives the freedom to dictate the solution function space with desired continuity and differentiability requirements. The DG based approach naturally ensures satisfaction of the entropy inequalities \cite{dafermos2005hyperbolic} and asymptotic consistency with the zero-viscosity limit without the need for any extra penalties in the objective function. 

In summary, the main technical contributions of this research are as follows: 
\begin{enumerate}

\item We propose a novel discontinuous Galerkin based deep learning framework to solve nonlinear PDEs with discontinuous solutions. The framework has the capability to predict solutions that may develop shocks or discontinuities in finite time without requiring any \textit{a-priori} information about the location of the discontinuities.

% with sharp gradients or discontinuous solutions.

% obtain weak solutions to hyperbolic PDEs. 

% \item The framework uses discontinuous Galerkin finite element approach to enforce physics-based constraints on the solution.

\item The use of DG approach easily allows to dictate continuity requirements of the function space of the outputs of neural network. This approach allows us to construct a function of certain regularity (continuity and differentiability) in the whole domain, as opposed to obtaining a compositional function of certain differentiability, giving the proposed approach a unique advantage.

\item Based on the DG-FEM discretization technique, the framework has the capability to capture sharp jumps or discontinuities in solution. This is achieved by augmenting the physical mesh with ghost elements as discussed later in Section \ref{sec:methodology}.

\item We leverage the weak form of the governing equations to reduce the regularity requirements on the solution. The weak form is compared against the predefined basis functions from the test function space to obtain the physics-based loss (objective function). Convolution operations are employed to numerically approximate the integrals in the weak form.

%  the composite loss is obtained from the weak formulation of the governing equations.

% which is identical to the DG-FEM discretization technique.

\item The application of initial conditions and Dirichlet/Neumann boundary conditions is straightforward and does not result in additional terms in the composite loss function. Essential/Dirichlet boundary conditions are exactly accounted for by the framework. The natural boundary conditions become part of the weak form akin to its treatment in FEM.

\item We test the performance of our framework by using it to approximate discontinuous function and solutions to hyperbolic PDEs including advection and Burgers equation. 
\end{enumerate}

% \cite{arora2019computational, arora2020dislocation, arora2020finite, arora2020unification}
% \cite{joshi2020equilibrium}
% \cite{arora2022mechanics}
% \cite{arora2021machine, arora2022physrnet, arora2022physics}
% \cite{aroraspatio}
% From leveque

% gradient enhance \cite{yu2022gradient}

% owning to the difficulty in resolving the shocks and discontinuities, 

% Although, there has been a flurry of research involving application of physics-informed ML models to various scientific fields

% Current network architectures share some of the limitations of classical numerical discretization schemes when applied to non-linear differential equations in continuum mechanics. A paradigmatic example is the solution of hyperbolic conservation laws that develop highly localized nonlinear shock waves. Learning solutions of PDEs with dominant hyperbolic character is a challenge for current PINN approaches, which rely, like most grid-based numerical schemes, on adding artificial dissipation. 

In the first part of this two-part treatise, we focus on presenting the details of the framework and then use the framework to compute solutions to scalar nonlinear hyperbolic PDEs in one dimensional space as a proof of concept.

\textbf{Organization:} The remainder of this paper is organized as follows:
A brief review of prior work on using PINNs to solve hyperbolic PDEs is discussed in Section \ref{sec:prior_works}.  Section \ref{sec:background} recalls the governing equations of a general hyperbolic PDE in its conservative form along with presenting the mathematical formulation. The details of the framework architecture, data setup, and physics-based loss function are presented in Section \ref{sec:methodology}. Section \ref{sec:results} presents the results for the numerical experiments performed in this work and demonstrates the superiority of the framework for approximating discontinuous functions along with solving hyperbolic PDEs such as advection and Burgers' equations. Finally, conclusion of the present work and outlook on future research are presented in Section \ref{sec:conclusion}.

\section{Prior Works}\label{sec:prior_works}

As presented in the seminal works of Raissi et al. \cite{raissi2017physicsI}, PINNs can be  be classified into two distinct classes of algorithms, namely continuous time and discrete time models. The former has drawn tremendous interest from the scientific community and has been shown to solve nonlinear PDEs in a wide variety of applications. However, these continuous time PINN models pose  serious limitation in that they require a large number of collocation points in the spatio-temporal domain to enforce physics-based constraints, rendering the training prohibitively expensive. Moreover, it is also a challenge for these models to predict the solutions to PDEs with advection dominant character wherein the solutions develop discontinuities in finite time even if the initial conditions are smooth \cite{dafermos2005hyperbolic}. Development of techniques that address this issue is therefore an active area of research in the PINN community.

Several recent works focus on the enhancement of shock capturing ability of PINNs for application to hyperbolic PDEs.  Jagtap et. al  \cite{jagtap2020adaptive} proposes adaptive activation functions in  PINN based models to facilitate approximation of discontinuous solutions to nonlinear PDEs. The approach adds a scalable hyper-parameter in the activation function which is optimized to improve the convergence rate and solution accuracy. Patel et. al \cite{patel2022thermodynamically} proposes control volume PINNS (cvPINNS) that adopts a least squares space-time control volume scheme to reduce solution regularity requirements for hyperbolic equations. Their approach leads to   introduction of a number of additional penalties in the loss to satisfy entropy inequality and total variation diminishing property on the solution. A Residual-based Adaptive Refinement (RAR) method has been proposed in  \cite{lu2021deepxde} to adaptively sample the collocation points near the discontinuity to improve the training efficiency of PINNs. Similar to the RAR method, \cite{mao2020physics} used clustered collocation points around sharp gradients to improve the solution accuracy to one-dimensional Euler equation with a moving contact discontinuity and a two-dimensional steady state problem with an oblique shock.

Several other works \cite{jagtap2020conservative,jagtap2021extended, dwivedi2019distributed}  divide the original computational domain into smaller sub-domains in which completely different neural networks can be employed while enforcing interfacial constraints at the interface of each sub-domain.  Lv et al \cite{lv2021hybrid} proposes hybrid PINNs that incorporates a heuristic based discontinuity indicator into the neural network to distinguish the non-smooth scales from the smooth regions. To compute the derivatives, automatic differentiation is used in smooth regions while the computationally expensive fifth-order weighted essentially non-oscillatory  (WENO) scheme is adopted to compute the derivatives in the vicinity of discontinuities (non-smooth scales). 
Another approach is highlighted in \cite{liu2022discontinuity} that weakens the strong form of the equations near the discontinuities by adaptively choosing a gradient-weight locally at each residual point. Coutinho and coworkers \cite{coutinho2022physics} introduced a variety of adaptive methods to automatically tune the dissipation term and studied its effectiveness on an  inviscid Burgers’ and Buckley–Leverett equations. Ryck et. al \cite{de2022wpinns} proposed  \textit{weak} PINNS based on approximating the solution of a min-max optimization problem for a residual, defined in terms of Kruzkhov entropies. 

Rodriguez et. al \cite{rodriguez2022physics} proposed another methodology called physics-informed attention-based neural networks (PIANNs) as a combination of recurrent neural networks and attention mechanisms to approximate sharp shocks in the PDE solutions. The approach requires a gated recurrent unit at each spatial location making it computationally intractable in higher dimensions. Moreover, the approach seems unable to perfectly propagate a strict discontinuity and smoothens the solution (see \cite[Figure 3]{rodriguez2022physics}). Xiong et. al  \cite{xiong2020roenets} introduced RoeNets to predict the discontinuous solutions to hyperbolic conservation laws. However, the approach is data-driven (requires training data) and therefore is uninformed of any physical insights based on the governing laws of the system. Moreover, the approach is not amenable to unstructured grids as the training data from numerical Roe solver uses a structured grid.

A prior attempt on using DG within a feed forward neural network was attempted in \cite{chen2021deep}. However, the approach presented therein suffers from two disadvantages: a) It requires the use of multiple neural networks when using higher-order polynomial interpolation in space. b) The approach fails to obtain second-order accuracy for a linear conservation law when using second order accurate schemes in space and time.

% \textcolor{red}{POINTWISE instead of feed forward or realte to introduction.}

% The approach also fails to demonstrate h-convergence, i.e. the errors failed to reduce when decreasing the mesh sizes. \textcolor{red}{explain more}.

% In summary, all the current approaches require one or more of the following techniques to learn the solutions of nonlinear PDEs with dominant hyperbolic character
% \begin{itemize}
%     \item a priori knowledge of the solution or location of discontinuities 
%     \item Heuristics based smoothness-indicator to distinguish smooth regions from non-smooth regions
%     \item Apply gradient-weight to governing equations to weakens the residual in the regions of discontinuity
%     \item Introduce additional penalizations for total variation diminishing constraint, satisfaction of entropy inequality, or interfacial constraints.
%     \item use of multiple neural networks while breaking the domain into multiple smaller sub-domains.
% \end{itemize}

% Therefore, we o

% However, despite these recent improvements, deep learning based approaches face serious challenges when applied to learn the solutions of nonlinear PDEs with dominant hyperbolic character. 

\section{Mathematical Preliminaries}\label{sec:background}

\subsection{Governing equations}\label{sec:gen_gov_eq}

% Since the PDEs with higher order spatial derivatives can be written as a system of first order equations \cite{bassi1997high}, it is sufficient to illustrate the idea for first-order hyperbolic equations.

A general first-order hyperbolic PDE in conservative form is given as
\begin{align}
\begin{split}
\dot{\bm{u}} + \nabla\cdot\bm{\mathcal{F}}(\bm{u}(\bm{x}, t), \bm{x}, t) = \bm{\mathcal{G}}(\bm{u}(\bm{x}, t), \bm{x}, t) ~~~~ \textrm{in}~~ \Omega\times\Sigma,
\label{eq:pde_cons}
\end{split}
\end{align}
subjected to the initial and boundary conditions
\begin{align}
\begin{split}
    \bm{u}(t=0, \bm{x}) = \bm{u}_0(\bm{x}),\\
    \bm{\mathcal{B}}(\bm{u};~ \bm{x}\in\partial\bm{\Omega}) = \bm{0}.
    \label{eq:dynamics_icbc}
\end{split}
 \end{align}
In equation \ref{eq:pde_cons}, $\Omega \subset \mathbb{R}^d$, $d\in \mathbb{N}$ denotes the spatial domain of the system and $\Sigma \subset \mathbb{R}^+$ denotes the temporal domain. The classical solution of the above PDE is a vector-valued differentiable solution $\bm{u}: \Omega\times\Sigma \rightarrow \mathbb{R}^m$, $m\in \mathbb{N}$ that satisfies the PDE at every point $(\bm{x},t)\in \Omega\times \Sigma$. $\dot{\bm{u}}$ denotes the time derivative of $\bm{u}$. The matrix-valued function $\bm{\mathcal{F}}:\mathbb{R}^m\times \Omega\times\Sigma\rightarrow\mathbb{R}^{m\times d}$ is referred to as the flux function. $\bm{\mathcal{G}}: \mathbb{R}^m\times \Omega \times \Sigma$ is any nonlinear vector-valued function of its arguments. The boundary of the spatial domain is denoted by $\partial\bm\Omega$ and $\bm{\mathcal{B}}$ denotes the boundary operator on $\bm{u}$. For a given problem, there can be
multiple boundary operators defined on different part of the boundary $\partial\Omega$.

% In its non-conservative form, the hyperbolic PDE can be re-written as:
% \begin{align}
% \begin{split}
%     \dot{\bm{u}} + \mathcal{B}(\bm{u}(\bm{x}, t), \bm{x}, t) : \nabla\bm{u} + \mathcal{C}(\bm{u}(\bm{x}, t), \bm{x}, t) = \bm{0}~~~~ \textrm{in}~~ \Omega\times\Sigma,
%     \label{eq:pde_non_cons}
% \end{split}
% \end{align}
% where $\mathcal{B}: \mathbb{R}^m\times \Omega\times\Sigma\rightarrow\mathbb{R}^{m\times d}$ and $\mathcal{C}: \mathbb{R}^m\times \Omega\times\Sigma\rightarrow\mathbb{R}^{m}$. 
% h are not differentiable, not even continuous, merely measurable and bounded.

% \begin{remark}
It is well established that the solutions to \eqref{eq:pde_cons} may develop shocks or discontinuities in finite time even if the the initial data is smooth \cite{leveque2002finite}. Therefore, the solutions to such system of PDEs are usually considered in the weak sense. However, these weak solutions to \eqref{eq:pde_cons} are not necessarily unique. Therefore, uniqueness is guaranteed by restricting attention to a class of (weak) solutions that satisfy entropy conditions. These details are beyond the scope of this work and the reader is referred to \cite{dafermos2005hyperbolic, leveque2002finite} and the references therein for a detailed overview of the theory of hyperbolic PDEs.
% \end{remark}

% Cite 38 from Discontinuous Galerkin Methods: General Approach and Stability.
\subsection{Discontinuous Galerkin Method}
\label{sec:dgmethod}

Discontinuous Galerkin (DG) methods \cite{reed1973triangular} are a special class of finite element methods that use completely discontinuous basis functions. This discontinuity of the basis function allows DG-FEM to have benefits that are not shared by typical finite element methods such as a) amenable to arbitrary triangulation with handing nodes b) p-adaptivity: complete freedom in choosing the order of basis functions in each element independent of that in the neighbours d) h-adaptivity: easily amenable to unstructured grids c) embarrassingly high parallel efficiency.  

% The $ith$ component of the vector valued solution $\bm{u} \in \mathcal{R}^m$ is denoted by $u_i$ ($1 \le i \le m$).
Next, we present key ideas of the discontinuous Galerkin FEM that forms the main physics engine behind the proposed framework.  The physical domain $\Omega$ is approximated by a space filling triangulation $\mathcal{T}$  with minimum grid size $h$ composed of $K$ geometry-conforming non-overlapping elements, $\mathcal{D}^k$
\begin{align}
    \Omega \approx \Omega_h = \bigcup_{k=1}^{K} \mathcal{D}^k.
\end{align} On each of these elements $\mathcal{D}^k$, we locally express the solution  $\bm{u}(\bm{x}, t)$ as a polynomial of order $N$ using $N_p$ basis function polynomials as follows ($N=N_p-1$):
\begin{align}
    \bm{u}^k(\bm{x}, t) = \sum_{n=1}^{N_p} \hat{\bm{u}}^k(t)\, {\Phi^k_n(\bm{x})}, ~~   \bm{x} \in \mathcal{D}^k.
    \label{eq:local_approx}
\end{align}
In equation \eqref{eq:local_approx}, $\Phi^k_n , n = 1,2, ..., N_p$ denotes the local polynomial basis on element $\mathcal{D}^k$ and $\hat{\bm{u}}^k$ denotes the nodal unknowns (expansion coefficients) for the  solution on the element $\mathcal{D}^k$.
The global solution $\bm{u}(\bm{x}, t)$ is then approximated by the direct sum of these N-th order local polynomial approximations $\bm{u}^k$
\begin{align}
    \bm{u}_h(\bm{x}, t) = \bigoplus_{k=1}^{K} \bm{u}^k(\bm{x}, t).
\end{align} We define a finite element space $\bm{U}_h$ as the space of all piecewise polynomial functions, continuous inside the elements $\mathcal{D}^k$ and (possibly) discontinuous on inter-element boundaries, defined on $\Omega_h$ that satisfy Dirichlet boundary conditions. $\bm{V}_h$ denotes the test function space  given by $\bm{V}_h = \{\bm{u}_h^1 - \bm{u}_h^2,  ~~\forall ~~\bm{u}_h^1, \bm{u}_h^2 \in \bm{U}_h\}$.  The discontinuous Galerkin formulation for the hyperbolic conservation law presented in  Equation  \eqref{eq:pde_cons} is then obtained from the following two steps: 

\textbf{Step 1}: The discontinuous Galerkin space discretization:
The PDE is first discretized in space using the discontinuous Galerkin method. A discontinuous approximate solution $\bm{u}_h \in \bm{U}_h$ is sought such that for all test functions $\delta \bm{u}_h \in \bm{V}_h$
\begin{align}
\begin{split}
   \bm{\mathcal{R}}^k(\bm{u}_h) := & \int_{\mathcal{D}^k} \dot{\bm{u}}_h \,\delta\bm{u}_h\,dV - \int_{\mathcal{D}^k} \bm{\mathcal{F}}: \nabla \delta\bm{ u}_h dV \\ & \quad +  \int_{\partial{\mathcal{D}^k}} \hat{\bm{\mathcal{F}}}\bm{n}^k\cdot\delta\bm{u}_h dA - \int_{\mathcal{D}^k} \bm{\mathcal{G}} \delta\bm{u}_h \,dV = 0.
   \label{eq:pde_weak}
\end{split}
\end{align}
This corresponds to a system of $N_p$ equations in each element $\mathcal{D}^k$. Therefore, the residual of the system $\bm{\mathcal{R}}$ is an array of size equal to total number of equations. In the above, $\bm{n}^k$ denotes the unit outward normal to the element boundary $\partial\mathcal{D}^k$  and $\hat{\bm{\mathcal{F}}}$ denotes the numerical flux which satisfies the Lipschitz continuity, consistency, and monotonicity conditions. $\hat{\bm{\mathcal{F}}}$ is a single valued function defined at the element interfaces and in general depends on the values of the numerical solution  from both sides of the interface. In this work, we use the well known Lax-Friedrichs flux for each equation in the system of conservation laws.

% as follows:

%  \begin{align}
%      \bm{\mathcal{F}}(\bm{u}^-, \bm{u}^+) =  \begin{cases}
%          \bm{u}^- & \bm{\mathcal{F}'}\cdot\bm{n}^k \ge 0\\
%          \bm{u}^+ & \bm{\mathcal{F}'}\cdot\bm{n}^k < 0,
%      \end{cases}
%  \end{align} where $\bm{\mathcal{F}'}= \frac{\partial \bm{\mathcal{F}}}{\partial \bm{u}}$

% \begin{align}
%     \hat{\bm{\mathcal{F}}}_i (\bm{u}^-, \bm{u}^+) =  \frac{1}{2} \left( \bm{\mathcal{F}}_i(\bm{u}^-) + \bm{\mathcal{F}_i}(\bm{u}^+) - \bm{A} \llbracket\bm{u}\rrbracket\right),
% \end{align}
%  where $\bm{A}$ denotes the vector of maximum eigenvalues of the Flux Jacobian $(\frac{\partial \bm{\mathcal{F}}_i}{\partial \bm{u}_i})$
 
% The above technique inherits all the merits of the usual finite element method. Complex geometries can be represented accurately and boundary conditions can be easily dealt with. \textcolor{red}{WRITE ABOUT FLUX from RKDG review last 174}

\textbf{Step 2}: The Range-Kutta time discretization: The above system of ordinary differential equations can then be discretized in time by following any of the explicit high-order accurate Runge–Kutta (RK) methods. In this work, we use the second order two stage Strong Stability Preserving Range-Kutta (SSP-RK) scheme \cite{gottlieb1998total, gottlieb2001strong}. The scheme for any intial value problem $\dot{\bm{y}} = \bm{\mathcal{L}}(t, \bm{y})$ is given as follows:
\begin{align*}
    \bm{v}^1 &= \bm{y}^t +\Delta t \, \bm{\mathcal{L}}(t, \bm{y}^t) ,\\
    \bm{y}^{t+\Delta t} &= \bm{y}^{t} + \dfrac{\Delta t}{2} \left(\bm{\mathcal{L}}(t+\Delta t, \bm{v}^1) + \bm{\mathcal{L}}(t, \bm{y}^t)\right).
\end{align*}

% DG - CD5 page 20/507

\begin{figure*}[t]
\centering
\begin{subfigure}{.95\textwidth}
  \includegraphics[width=\linewidth]{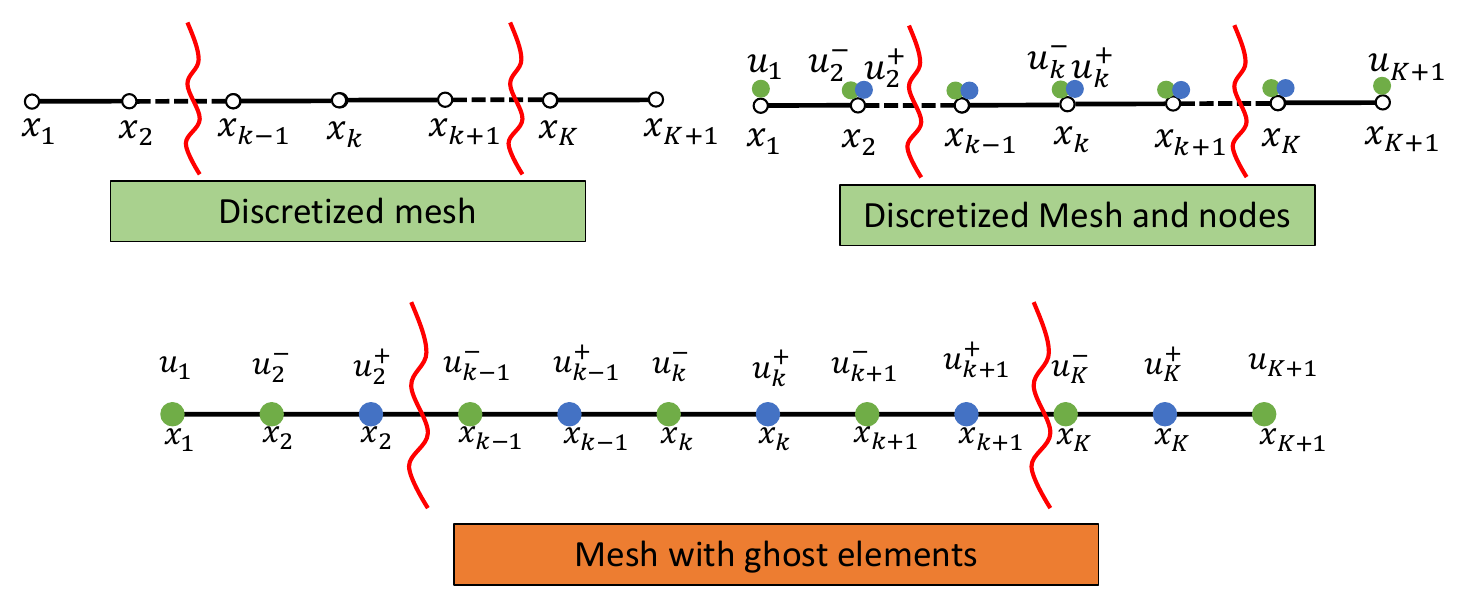}
\end{subfigure}
  \caption{The Figure on top left shows the discretized mesh for a 1-dimensional domain. The top right figure shows the nodal unknowns for the discontinuous Galerkin based discretization approach. The bottom figure shows the physical mesh augmented with ghost elements. This allows the framework to predict discontinuous solutions.}
  \label{fig:mesh}
\end{figure*}

\section{Methodology}
\label{sec:methodology}

In this section, we develop the setting of the discontinuous Galerkin based deep learning framework proposed in this work.  Taking motivation from the traditional finite element method (FEM) that solves for solution values at points (nodes) in the discretized domain, the framework is similarly used to predict nodal values.  The nodal values are then interpolated by using piecewise continuous basis functions to yield a globally defined solution field. This approach allows us to construct a function of certain regularity (continuity and differentiability) in the whole domain, as opposed to obtaining a compositional function of certain differentiability, making the proposed approach uniquely advantageous. The polynomial order and the continuity of the interpolating basis functions can be further leveraged to generate nonlinear solutions with desired continuity requirements.

Figure \ref{fig:mesh} (top left) shows a discretized mesh in one dimension in which each element shares the nodes with a neighbouring element. Discontinuous Galerkin based technique adds additional degrees of freedom at each shared node along the faces of the elements as shown in Fig.\ref{fig:mesh} (top right). In a similar spirit, we augment the physical mesh with ghost elements to allow the neural network to model jump or discontinuities in solutions as shown in Fig.\ref{fig:mesh} (bottom).

% The use of DG based approach easily allows to dictate continuity requirements of the function space of the outputs of neural network.  Based on the DG-FEM discretization technique, the framework has the capability to capture sharp jumps or discontinuities in solution. This is achieved by augmenting the physical mesh with ghost elements.

Once the domain is discretized, we leverage the weak form of the governing equations to reduce the regularity requirements on the solution. The weak form is compared against the predefined basis functions from the test function space to obtain the physics-based loss (objective function).  The integration is performed using Gauss quadrature schemes and the convolution operations are used to numerically evaluate the integrals in the weak form.

% Section \ref{sec:network_arch} discusses the architecture of discontinuous Galerkin based deep-learning framework proposed in this work. 
% Section \ref{sec:loss} briefly outlines the physics-informed composite loss (objective function) used during the framework's training. 

% An integral part of this approach is the design of the loss function. The finite element based loss function is inspired by the Galerkin formulation of an elliptic PDE as well as the Rayleigh-Ritz method. 

% In this case, we actually construct a function of certain regularity in the domain variable x, as opposed to just
% assuming a function of certain differentiability at the collocation points.

% \textcolor{red}{WRITE ABOUT CNN VS POINTWISE APPROACH. INTEGRATION CONVOLUTION MESH with ghost. How FENEuNet is different/limitations. WEAK FORM. COMPARISON WITH VPINNS and hp PINNs}

We note a few similarities and several differences with a previous research  \cite{khara2021neufenet} that developed an FEM based neural architecture to solve PDEs (with continuous solutions). The approach presented therein relies on the  use of an energy functional that can be minimized to yield the solution (Rayleigh-Ritz approach). However, the current work focuses on integrating discontinuous Galerkin based approach in a deep-learning based framework and relies on the minimization of the residual of the weak form.

\subsection{Input and output for the framework}
The input to the framework consists of the value of the solution field  $\bm{u}^t$ at any time $t$. Based on the second order two stage Strong Stability Preserving Range-Kutta (SSP-RK) scheme \cite{gottlieb1998total, gottlieb2001strong}, the solution at the next time instant $\bm{u}^{t+\Delta t}$ is given as
\begin{align}
\begin{split}
&\bm{v}^{1} = \bm{u}^t + \Delta t \, \bm{\mathcal{NN}}(\bm{u}^t;\theta),\\
\bm{u}^{t+\Delta t} &= \bm{u}^t + \frac{\Delta t}{2} \, \left( \bm{\mathcal{NN}}(\bm{u}^t;\theta) + \bm{\mathcal{NN}}(\bm{v}^1;\theta) \right),
\label{eq:rke}
\end{split}
\end{align}
where $\bm{\mathcal{NN}}(\cdot; \theta)$ denotes the framework outputs. The solution field $\bm{u}^{t+\Delta t}$ becomes the input to the framework at time instant $t+\Delta t$. $\theta$ denotes the set of weights and biases of the network.

\begin{figure*}[t]
\centering
\begin{subfigure}{.95\textwidth}
  \includegraphics[width=\linewidth]{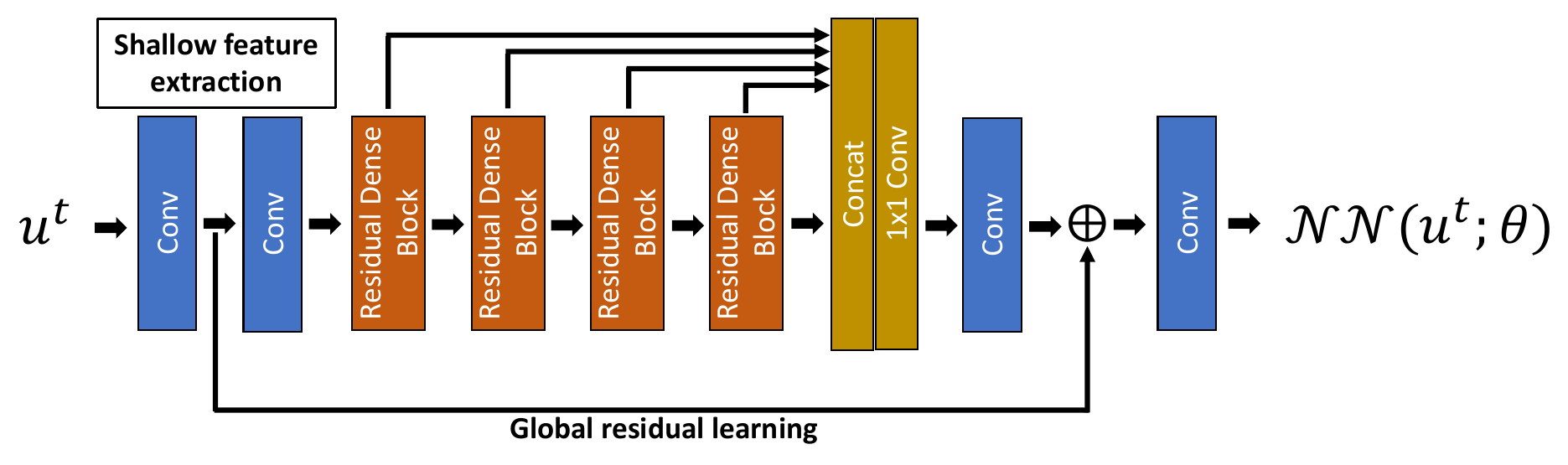}
\end{subfigure}
  \caption{The Figure shows the architecture of the proposed framework.}
  \label{fig:arch}
\end{figure*}

\subsection{Architecture}

In this work, we implement the framework as a variant of the Residual Dense Network (RDN) originally proposed in \cite{zhang2018residual}. In the context of PINNS, RDN has previously been shown to be successful in enhancing the spatial and temporal resolutions of coarse-scale (both in space and time) PDE solutions without requiring any labeled data \cite{arora2021machine,arora2022physrnet,arora2022spatio}. Figure \ref{fig:arch} shows the architecture of the proposed framework. The architecture consists of 
\begin{enumerate}
    \item \textbf{Shallow feature extraction}: Two convolutional layers are used to extract shallow features from the input state variable $u_i$ $(i = 0, 1, ..., T  -1)$, where $T$ denotes the total number of time steps. The output from the first convolutional layer is later reused below in step $4$.

    \item \textbf{Global feature fusion}: Residual dense blocks with ReLU activation functions are then stacked together for extracting  local dense features. The features extracted from all residual blocks are then concatenated together to exploit hierarchical features in a global way.

    \item  The concatenated hierarchical features are fed to a $1\times 1$ convolution layer to adaptively fuse a range of features with different levels followed by another convolution layer to further extract features for global residual learning.

    % RDBs produce multi-level local dense features, which are further adap- tively fused to form FGF . After global residual learning, we obtain dense feature FDF .
    
    \item \textbf{Global residual learning}: The shallow features (from step $1$)  and the globally fused features (output from step $3$) are added together before the final output layer.
    
    \item In the end, another convolutional layer is added to produce the output with desired dimensions. %There is no activation function used in the output layer.
\end{enumerate}

\subsection{Objective function} \label{sec:loss}
% Most of the research involving the use of PINNs to approximate solutions to PDEs employ the strong form of the governing equations in the loss function. This requires a suitably chosen set of collocation points in the domain at which the loss is evaluated. However, more recently, based on the method of weighted residuals \cite{finlayson1966method}, new PINN based approaches have been developed  \cite{khodayi2020varnet, kharazmi2019variational, kharazmi2021hp}. These solvers 

The solution to the system of equations \eqref{eq:pde_cons} is obtained by solving the weak form \eqref{eq:pde_weak} such that the norm of the residual $||\bm{\mathcal{R}}||$ becomes smaller than an acceptable tolerance value. In other words, the solution can also be written as the minimizer of the norm of residual $||\bm{\mathcal{R}}||$ such that $\bm{u}_h = \arg\min_{\bm{u}_h\in V_h} ||\bm{\mathcal{R}}(\bm{u}_h)||$.
Therefore, the physics-informed objective function to be minimized during the training is given as follows:
% Since the entire training stage is unsupervised, the objective/loss function for the training is directly obtained from the weak from of DG formulation. The physics-informed objective function is then written as follows:
\begin{align}
\bm{u}_h(\theta) = \arg\min_{\theta}||\bm{\mathcal{R}}(\bm{u}_h(\theta))||_1,
	% \begin{split}
		% \mathcal{L} ~=~ \,  \underbrace{||\bm{\mathcal{R}(\bm{u(\bm{x}, t)},\bm{x}, t)}||_1}_{\text{weak form}},
  % \end{split}
\label{eq:general_loss}
\end{align} 
where $||\bm{A}||_1$ denotes the $L_1$ loss which is the mean absolute error (MAE) between each element in the quantity $\bm{A}$ and target $0$.

% \subsection{Implementation details}

\subsection{Boundary conditions}
The proposed approach naturally accounts for the boundary conditions (Dirichlet and Neumann) without the need to treat them as extra penalty terms in the composite loss thereby avoiding issues arising during multi-objective optimization \cite{wang2020understanding}.

% reducing requisite hyperparameters in the process.

\noindent\textbf{Dirichlet boundary conditions}: Similar to finite element method based approaches, the Dirichlet boundary conditions are satisfied exactly in the framework. Once the framework predicts the solution at the next time step $\bm{u}^{t+\Delta t}$, a simple post-processing step is applied to append the  known boundary values to the prediction. This process allows for exact imposition of Dirichlet boundary conditions without any additional penalty terms in the loss function. This process also eliminates the need to use distance functions (analytical or pre-trained models) to impose Dirichlet boundary conditions making the training process more interpretable.

% to the discretized grid based solution 

\noindent\textbf{Neumann boundary conditions}: The natural boundary conditions are included in the weak form of the PDE. Therefore, the natural boundary conditions are automatically satisfied in the weak sense at the discrete form akin to its treatment in finite element based techniques. 

% Since they become part of the weak form, the natural boundary conditions do not appear 

% This is done by simply ignoring the nodal value of $\mathcal{R}$ for nodes that are constrained by the Dirchlet boundary condition. 

\begin{figure*}[t]
\centering
\begin{subfigure}{.90\textwidth}
  \includegraphics[width=\linewidth]{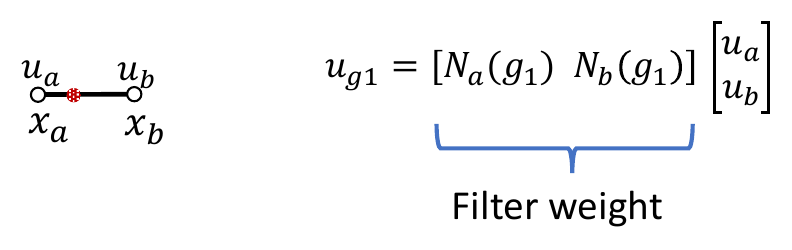}
\end{subfigure}
  \caption{Figure showing a physical element from the mesh in $1$-$d$ and a gauss point $g_1$ marked in red. $x_a$ and $x_b$ denote the nodal coordinates. $u_a$ and $u_b$ denote the value of the solution at those nodal points. The equation on the right shows the convolution filter weight used to evaluate the value of the solution $u$ at the Gauss point $g_1$ (assuming linear interpolation). The filter weight can be obtained similarly when using higher order interpolation.}
  \label{fig:gauss_integration}
\end{figure*}

\subsection{Integration and derivatives} \label{sec:int_and_deri}
Integration over the domain and boundaries in the finite element based approaches are simply calculated by evaluating the sum of the integration over individual elements interiors or elements boundaries
\begin{align}
\begin{split}
    \int_\Omega f dV = \sum_{i=1}^K \int_{D^k} f dV\\
    \int_{\partial\Omega} g dA = \sum_{i=1}^K \int_{\partial\mathcal{D}_o^k} g dA.
    \end{split}
\end{align}
where $D_o^k$ represents the exterior boundary for any element $D^k$. To numerically approximate the integrals, we use the Gaussian quadrature rule which is the weighted sum of integrand values at specified points within the domain of integration. We note that this evaluation of the integrand  values inside the elements can be done with the use of a convolution operation. Figure \ref{fig:gauss_integration} presents the visualization of the process of using a  convolution operation for evaluating the integrand values at an interior point. For each quadrature point, the convolution filter is essentially the values of the interpolating basis functions at that point assembled together in a matrix (array in one dimension).

\subsection{Training}

The framework is implemented and trained using PyTorch. We use Adam optimizer with an initial learning rate of $1\times10^{-3}$. As the training progresses, the learning rate is adjusted using  \texttt{ReduceLROnPlateau} scheduler. 

% with \texttt{patience} set to $40$. 

\section{Results and discussion}
\label{sec:results}

For all the numerical experiments, the physical domain is distributed into $128$ equally spaced elements along the $x$-direction and  $\Delta t = 0.004$ is used. The architecture of the framework uses $4$ residual dense blocks with $8$ layers in each block. The growth rate and the number of features are set to $32$ in the model architecture. We use the same architecture for all the experiments and do not excessively tune hyperparameters individually for each case. The experiments are performed on an NVIDIA Tesla V100 GPU card with 32 GB RAM. For ease of presentation, we restrict the remainder of the section to a $1$-d spatial domain with a Cartesian mesh, although the framework is easily generalizable to polyhedral meshes. We also use linear shape functions in each (physical) element and a 2-point Gauss quadrature scheme to evaluate the integrals in the weak form.

% Advection + Burgers: https://arxiv.org/pdf/1907.08967.pdf

% Variational Physics Informed Neural Networks: the Role of Quadratures and Test Functions
% The main differences with respect to the PINN are that the weak formulation of the PDE is exploited, the collocation points are replaced by test functions, and quadrature points are used to compute the integrals involved in the variational residuals. 

\subsection{Approximation of nonlinear function with a static discontinuity}
In this test, we use the framework to solve a differential equation with static discontinuity at $x = 0$ (i.e., the discontinuity stays fixed at $x=0$). The differential equation and the initial conditions are given as 
\begin{align*}
    \dot{u}(x,t) =
    \begin{cases}
     \cos(12x)       & \quad x\ge 0\\
     \sin(6x)  & \quad x < 0,
  \end{cases}
\end{align*}
\begin{align*}
    u(x, t=0) =
   \begin{cases}
    0.5\cos(12x)       & \quad x\ge 0\\
    0.2\sin(6x)  & \quad x < 0.
  \end{cases}
\end{align*}
The analytical solution is given as 
\begin{align*}
    u(x, t) =
   \begin{cases}
    (0.5+t)\cos(12x)       & \quad x\ge 0\\
    (0.2+t)\sin(6x)  & \quad x < 0.
  \end{cases}
\end{align*}
Here, the domain is [-2, 2].  Figure \ref{fig:result_dc} shows the solution obtained by the framework along with the analytical solution at three different time instants. We note that the obtained solution matches well with the analytical solution and successfully captures the discontinuity at $x=0$. As reflected from the Figure, the framework is also able to capture the high frequencies in the solution within $1000$ training epochs. The mean squared errors for the solutions presented in Fig. \ref{fig:result_dc} are $\approx 2\times 10^{-10}$. Therefore, we conclude that the framework successfully captures the discontinuity at $x=0$ with high accuracy without the need for any adaptive refinement or \textit{a-priori} knowledge of the location of the discontinuity.

\begin{figure*}[t]
\centering
\begin{subfigure}{1\linewidth}
  \includegraphics[width=\linewidth]{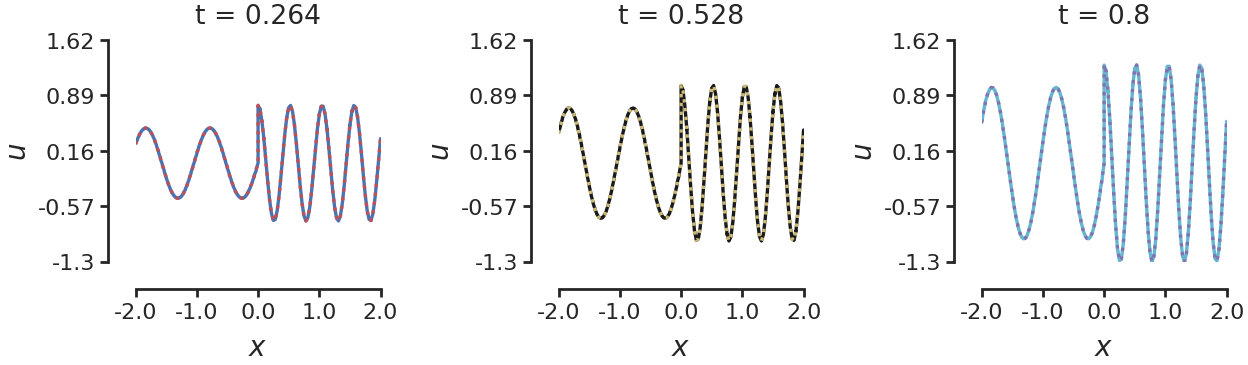}
% \label{fig:adv_dc}
\end{subfigure}%
  \caption{The figure shows the solution of an ordinary differential equation with static discontinuity (jump at $x=0$) at three different time instants. Solid lines represents the solution obtained from the framework and dashed lines represents the exact solution.}
\label{fig:result_dc}
\end{figure*}

\subsection{Advection equation}
\label{sec:adv}

In this section, we  consider a one-dimensional advection equation that has the following mathematical form
\begin{align}
    \frac{\partial u(x,t)}{\partial t} + v(x,t) \frac{\partial u(x,t)}{\partial x} = 0.
    \label{eq:adv1d}
\end{align}
We present the results for the following two cases: a) Smooth initial condition and b) Discontinuous initial condition.  For both the cases, we take the domain $x \in[-2, 2]$ and the velocity field to be constant, i.e.~$v(x,t) = 1$. We note that the traditional finite element method fails to converge when solving the advection equation. Therefore, a discontinuous Galerkin based discretization is an obvious choice.

\begin{figure*}[t]
\centering
\begin{subfigure}{\linewidth}
  \includegraphics[width=\linewidth]{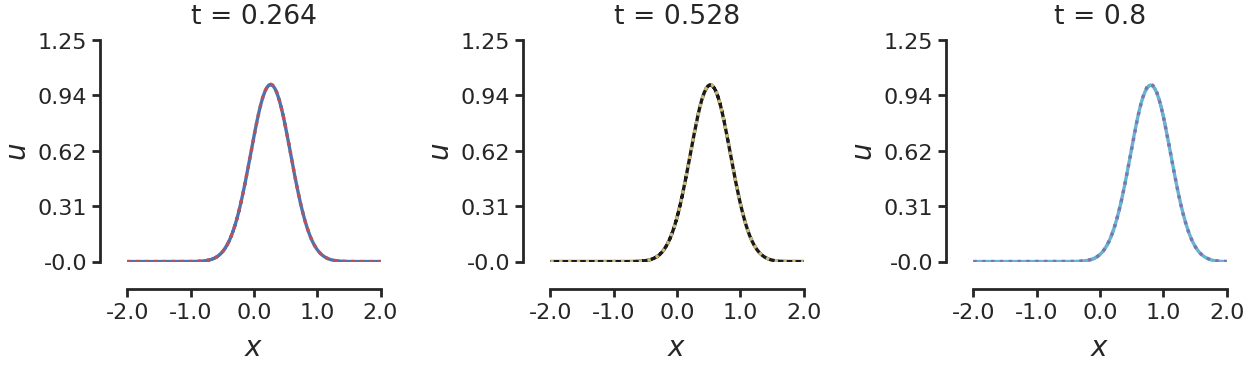}
% \caption{(solid line represents framework solution and dashed lines represents exact solution)}
  % \label{fig:error_model}
\end{subfigure}%
\caption{The figure shows the solution of the advection equation with smooth initial condition at three different time instants. Solid lines represents the solution obtained from the framework and dashed lines represents the exact solution.}
\label{fig:adv_exp}
\end{figure*}

\subsubsection{Smooth initial condition}
This section focuses on applying the framework to solve the advection equation with continuous initial conditions given by
\begin{align}
u(x, t=0) &= e^{-5x^2},~~~~\forall~x\in [-2,2].
% u(x, t=0) &= \sin(2\pi x), ~~~~\forall x\in [-1,1]
\end{align}
 The boundary condition at $x=-2$ is taken as $u(x=-2, t) = 0$.  The analytical solution for this problem is the traveling wave solution $u(x,t) = e^{-5(x-t)^2}$. 
Figure \ref{fig:adv_exp} shows the results obtained from the framework   along with the analytical solution at three different time instants. The framework is able to solve the advection equation with great accuracy and is able to match the analytical travelling wave solution. This verifies our implementation of the discontinuous Galerkin discretization within the convolutional neural network framework.

\begin{figure*}[t]
\centering
\begin{subfigure}{\textwidth}
  \includegraphics[width=\linewidth]{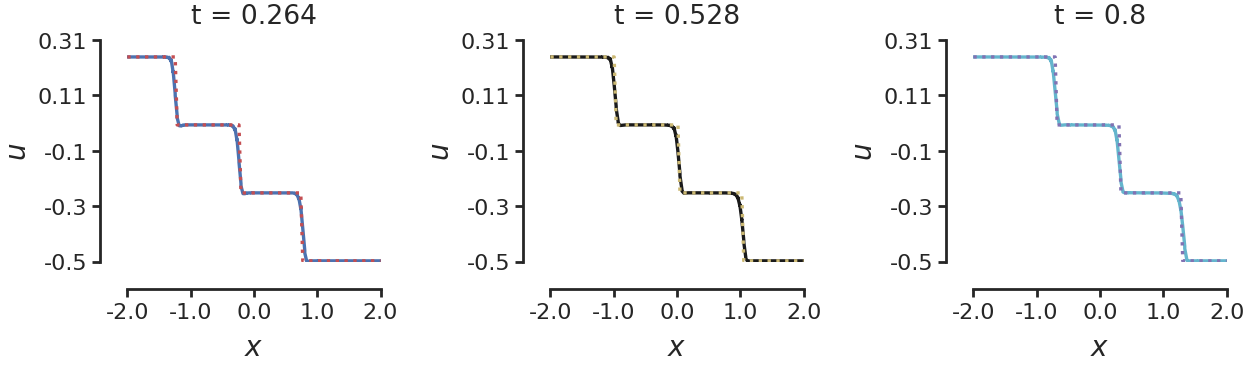}
  % \caption{(solid line represents framework solution and dashed lines represents exact solution)}
  % \label{fig:error_model}
\end{subfigure}%
\caption{The figure shows the solution of the advection equation with jumps in the initial condition at three different time instants. Solid lines represents the solution obtained from the framework and dashed lines represents the exact solution.}
\label{fig:adv_jump}
\end{figure*}

\subsubsection{Discontinuous initial condition}
This section focuses on applying the framework to solve the advection equation with discontinuous initial conditions. The initial and boundary conditions are given as 
\begin{align}
u(x, t=0) &= \begin{cases}
    0.25       & \quad x\le -1.5\\
    0.0       & \quad x> -1.5~\text{and}~x\le -0.5\\
    -0.25       & \quad x>-0.5~\text{and}~x\le 0.5\\
    -0.5       & \quad\text{otherwise}.
\end{cases}, ~~~~\forall~x\in [-2,2]\\
u(x=-2, t) &= 0.25.
\end{align} The analytical solution for this problem is the traveling wave solution given by 
\begin{align}
u(x, t) &= \begin{cases}
    0.25       & \quad x-t\le -1.5\\
    0.0       & \quad x-t> -1.5~\text{and}~x-t\le -0.5\\
    -0.25       & \quad x-t>-0.5~\text{and}~x-t\le 0.5\\
    -0.5       & \quad\text{otherwise}.
\end{cases}, ~~~~\forall~x\in [-2,2]
\end{align} We note that the location of the discontinuity (jump in solution) moves in space as the time progresses. For the conventional PINN based approaches, solving the advection equation with such an initial condition would necessitate the use of residual based adaptive refinement. The approach would also require a large amount of training data in the regions of sharp gradients. However, in the absence of any labeled data, such an information is usually not known a-priori. Therefore, traditional PINN based approaches face serious difficulties in capturing the sharp gradients in the solutions.

 Figure \ref{fig:adv_jump} shows the comparison of the solution obtained from the framework with the analytical solution at three different time instants. The framework is able to resolve the sharp jumps in solution at all times without any apriori knowledge of its location or use of adaptive refinement.

% Figure \ref{} shows the schematic of the body along with the imposed boundary conditions.

% An isotropic body is considered to be deforming under anti-plane strain conditions in two-dimensions. Under such conditions, the unknown components for displacement, velocity, and stress fields are $u_z$, $v_z$, and ($\sigma_{xz}, \sigma_{yz}$), respectively.

% HR ground truth data 
% This LR data is then used as input to the super-resolution framework.

\begin{figure*}[t]
\centering
\begin{subfigure}{\textwidth}
  \includegraphics[width=\linewidth]{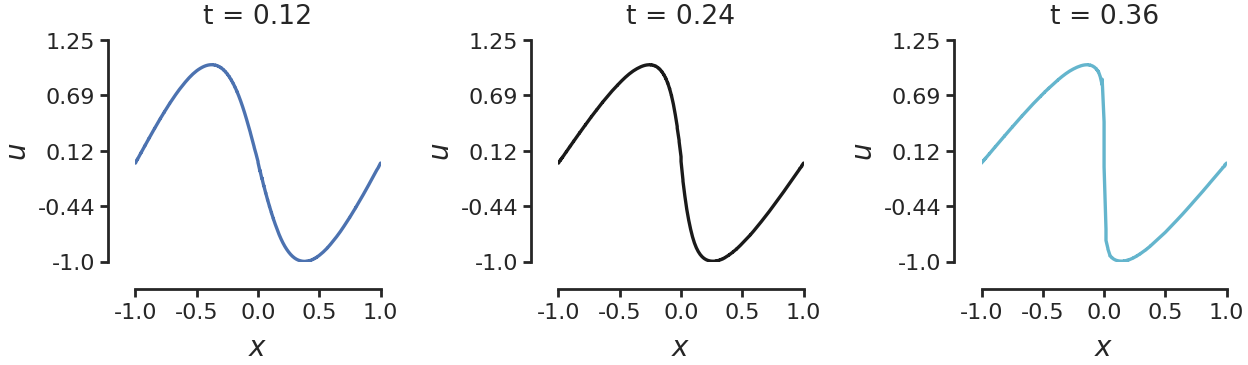}
  % \caption{(solid line represents framework solution and dashed lines represents exact solution)}
  % \label{fig:error_model}
\end{subfigure}%
\caption{The figure shows the solution of the inviscid Burgers equation with periodic boundaries and sinusoidal initial condition at three different time instants. Solid lines represents the solution obtained from the framework and dashed lines represents the exact solution.}
\label{fig:burgers}
\end{figure*}

\begin{table}[ht]

\begin{center}

\begin{tabular}{ |p{2cm}|p{2cm}|p{8cm}| }
% \begin{tabular}{ |m{1.5cm}|m{1.4cm}|m{8cm}|m{4cm}|  }
% \begin{tabular}{ | c | c | c | c |}
\hline
\multicolumn{3}{|c|}{Comparison with other works} \\
\hline
  \footnotesize{Collocation points} & \footnotesize{Reference} & Disadvantage \\
\hline
300 & \cite{lv2021hybrid}  & \beginmyitemize \item Requires computationally expensive fifth-order WENO finite difference scheme \cite[see Table 6]{borges2008improved}  \item Not amenable to unstructured grids \end{itemize}\\
\hline
10,000 & \cite{liu2022discontinuity}  & \beginmyitemize \item Requires Large number of collocation points \item  Effectiveness of the weight is yet to be demonstrated for general problems \end{itemize}\\
\hline
4,141 & \cite{dwivedi2019distributed}  & \beginmyitemize \item Breaks the domain into smaller smaller sub domains requires location of the discontinuity to be known a-priori \end{itemize} \\
\hline
-- & \cite{raissi2017physicsI, jagtap2020adaptive, liu2022discontinuity, dwivedi2019distributed, coutinho2022physics, patel2022thermodynamically, jagtap2020conservative}& Solve viscous Burgers equation \\
\hline

\end{tabular}
\end{center}
\caption{Table mentioning the other works that solve the inviscid Burgers equation and its disadvantages as compared with the proposed framework.}
\label{tab:burgers_comp}
\end{table}

\subsection{Burgers equation}

We apply our discontinuous Galerkin based deep-learning framework to solve an inviscid Burgers equation in this section. The inviscid Burgers equation is a nonlinear first order hyperbolic PDE that can develop shocks or discontinuities even if the initial data is smooth. Therefore, it has become a standard benchmark to demonstrate the shock capturing ability of any numerical scheme including PINN based approaches. The one dimensional inviscid Burgers equation is given as
\begin{align}
    \frac{\partial u(x,t)}{\partial t} + u(x,t)\frac{\partial u(x,t)}{\partial x} = 0, ~~~~~~ \forall~x \in [-1,1].
\end{align} We use a sinusoidal initial condition with periodic boundaries in this work, i.e.
\begin{align}
    u(x, t=0) = -\sin(\pi x )\\
    u(x=-1, t) = u(x=1,t) = 0.
\end{align} The exact solution to the Burgers equation for these set of initial and boundary conditions develops a shock at $x=0$. Table \ref{tab:burgers_comp} presents brief details of some of the other works that solve inviscid Burgers equation using physics informed neural networks. 

Figure \ref{fig:burgers} shows the solution obtained from the framework at three different time instants. We notice that the framework is able to solve the Burgers equation with great accuracy without any need for adaptive refinement or \textit{a-priori} knowledge of the location of the discontinuity. The framework is successfully able to approximate the solution to the inviscid Burgers equation which involves self steepening into a sharp profile at $x=0$ even when the initial condition is a smooth sinusoidal wave.

\section{Conclusion}
% \textcolor{red}{FINISH CONCLUSION}
% \textcolor{red}{use h convergence}
\label{sec:conclusion}

% The future works of this research will focus on applying the framework to solve PDEs in higher dimensions and presenting a more comprehensive evaluation of framework's performance against other PINN based approaches.

% Since DG method is a class of finite element method, the framework can easily be generalized to solve multi-dimensional problems in the same fashion as in the one-dimensional case presented here. The framework also enables exact enforcement of complex boundary conditions (both Dirichlet and Flux). Moreover, the extension of the framework to arbitrary triangulations can be easily be done via coordinate transformation between the physical and reference domains \cite{gao2020phygeonet}. 

% the framework is  is easy to handle complicated geometry and boundary conditions.

% In this work, we presented an unsupervised discontinuous Galerkin based deep learning framework to solve hyperbolic PDEs. The framework is a first in the literature that has the following capabilities:

This research focused on the development of deep-learning based framework that can predict discontinuous solutions to nonlinear PDEs without any \textit{a-priori} knowledge of the solution or the location of the discontinuities.  The work takes motivation from the traditional discontinuous Galerkin based finite element method (FEM) that solves for solution values at points (nodes) in the discretized domain. These nodal values are interpolated by using piecewise continuous basis functions to yield a globally defined solution field. The polynomial order and the continuity of the interpolating basis functions is leveraged to generate nonlinear solutions with desired continuity requirements. The physical mesh is augmented with ghost elements to allow for jump in solutions and integration is performed using Gauss point quadrature schemes. Built on the rigorous mathematical foundations of the discontinuous Galerkin method, the framework naturally handles imposition of boundary conditions (Neumann/Dirichlet), entropy conditions, and regularity requirements. Several numerical experiments and validation with analytical solutions demonstrate the accuracy, robustness, and effectiveness of the proposed framework.  

% \textcolor{red}{mesh ghost elements. Integration in weak form and Gauss point integration is implemented with a CNN framework NO ABSTRACT}

% \textcolor{red}{mesh ghost elements. Integration in weak form and Gauss point integration is implemented with a CNN framework}

% \begin{itemize}

% \item Faithfully solves hyperbolic PDEs by and the solutions are weak solutions

% \item The objective function is the true weak form of the PDE

% \item The 

% \item Exactly satisfies the given initial and boundary conditions.

% \item Enables spatial and temporal upscaling (resolution enhancement) of coarse-grained solutions to spatio-temporal PDEs while ensuring that the (upscaled) outputs satisfy the governing laws of the system.

% \item Easily allows imposition of any additional constraints (PDE or algebraic) in the framework.

% \item It is easily generalizable to non-rectangular domains by using  coordinate transformation between physical and reference domains as outlined in \cite{gao2020phygeonet}.

% \item The framework is highly scalable to clusters with multi-gpu nodes using PyTorch's \texttt{DistributedDataParallel}  \cite{li2020pytorch} functionality.

% for workloads that require substantial computational resources.

% \end{itemize}

The current work also paves the way for the following extensions in the future works:
\begin{itemize}
   
\item  Given the generality and universality of the proposed formalism, the framework can be easily extended to multi-dimensions and applied to study a wide range of problems in fluid in solid mechanics. Furthermore,  studying the effects of high-order piecewise polynomials and slope-limiters on model accuracy and convergence is another interesting research area to pursue.

\item The future research will also study the h-convergence (mesh convergence) and p-convergence (convergence based on basis order) of the framework analogous to the convergence analysis of the finite element based solution to PDEs.

 \item Although the current work focuses on the forward problems, future work would also focus on applying the framework to study  inverse problems and parametric PDEs where PDEs are defined by a family of parametrized physical properties or boundary conditions. 

\item The future works of this research will focus on presenting a more comprehensive evaluation of framework's performance against other PINN based approaches while studying nonlinear problems with moving shocks/discontinuities in higher dimensions. 

\item The extension of the framework to arbitrary triangulations can be done via coordinate transformation between the physical and reference domains \cite{gao2020phygeonet}. However, the use of Graph Neural Networks (GNN) to account for the mesh structural information and learning over unstructured grids and complex topological structures is another interesting avenue to pursue.

\item The use of Local Discontinuous Galerkin (LDG) method to solve equations with higher order derivatives is a work in progress.

\end{itemize}

\section*{Acknowledgment}
The work was conceptualized during the authors' time at Carnegie Mellon University (CMU). The author thank Dr. Ankit Shrivastava, research associate at Sandia National Laboratories, for useful discussions and his comments on the manuscript.

% \section*{Footnotes}

% \section*{References}

\clearpage
\newpage

\bibliographystyle{unsrt} % We choose the "plain" reference style
\bibliography{main} % Entries are in the refs.bib file

\end{document}